\documentclass[conference]{IEEEtran}
\IEEEoverridecommandlockouts
\usepackage{cite}
\usepackage{amsmath,amssymb,amsfonts,bm}
\usepackage{algorithmic}
\usepackage[inline]{enumitem}
\usepackage{booktabs}
\usepackage{graphicx}
\ifCLASSOPTIONcompsoc
    \usepackage[caption=false,font=footnotesize,labelfont=sf,textfont=sf]{subfig}
\else
    \usepackage[caption=false,font=footnotesize]{subfig}
\fi
\usepackage{textcomp}
\usepackage{xcolor}
\def\BibTeX{{\rm B\kern-.05em{\sc i\kern-.025em b}\kern-.08em
    T\kern-.1667em\lower.7ex\hbox{E}\kern-.125emX}}
\newcommand{\heading}[1]{\multicolumn{1}{c}{#1}}    

\usepackage{tikz}
\usetikzlibrary{
    calc,
    chains,
    shapes, shapes.geometric,
    arrows, arrows.meta
}
\definecolor{set1-darkred}{RGB}{215,48,39}
\definecolor{set1-lightred}{RGB}{252,141,89}
\definecolor{set1-verylightred}{RGB}{254,224,144}
\definecolor{set1-verylightblue}{RGB}{224,243,248}
\definecolor{set1-lightblue}{RGB}{145,191,219}
\definecolor{set1-darkblue}{RGB}{69,117,180}

\newcommand{\resfigscale}{0.48}

\begin{document}

\onecolumn

\noindent \textcopyright{} 2019 IEEE. Personal use of this material is permitted. Permission from IEEE must be obtained for all
other uses, in any current or future media, including reprinting/republishing this material for advertising or
promotional purposes, creating new collective works, for resale or redistribution to servers or lists, or reuse
of any copyrighted component of this work in other works.

\twocolumn

\title{%
On the Estimation of Complex Circuits Functional Failure Rate by Machine Learning Techniques
\thanks{This work was supported by the RESCUE project which has received funding from the European Union's Horizon 2020 research and innovation programme under the Marie Sklodowska-Curie grant agreement No. 722325.}
}

\author{%
\IEEEauthorblockN{%
  Thomas Lange\IEEEauthorrefmark{1}\IEEEauthorrefmark{2},
  Aneesh Balakrishnan\IEEEauthorrefmark{1}\IEEEauthorrefmark{3},
  Maximilien Glorieux\IEEEauthorrefmark{1},
  Dan Alexandrescu\IEEEauthorrefmark{1},
  Luca Sterpone\IEEEauthorrefmark{2}%
}
\IEEEauthorblockA{%
  \IEEEauthorrefmark{1}\textit{iRoC Technologies}, Grenoble, France \\
  \IEEEauthorrefmark{2}\textit{Dipartimento di Informatica e Automatica, Politecnico di Torino}, Torino, Italy \\
  \IEEEauthorrefmark{3}\textit{Department of Computer Systems, Tallinn University of Technology}, Tallinn, Estonia \\
  \{thomas.lange, aneesh.balakrishnan, maximilien.glorieux, dan.alexandrescu\}@iroctech.com \qquad
  luca.sterpone@polito.it}
}

\maketitle

\begin{abstract}

De-Rating or Vulnerability Factors are a major feature of failure analysis efforts mandated by today's Functional Safety requirements. Determining the Functional De-Rating of sequential logic cells typically requires computationally intensive fault-injection simulation campaigns. In this paper a new approach is proposed which uses Machine Learning to estimate the Functional De-Rating of individual flip-flops and thus, optimising and enhancing fault injection efforts. Therefore, first, a set of per-instance features is described and extracted through an analysis approach combining static elements (cell properties, circuit structure, synthesis attributes) and dynamic elements (signal activity). Second, reference data is obtained through first-principles fault simulation approaches. Finally, one part of the reference dataset is used to train the Machine Learning algorithm and the remaining is used to validate and benchmark the accuracy of the trained tool. The intended goal is to obtain a trained model able to provide accurate per-instance Functional De-Rating data for the full list of circuit instances, an objective that is difficult to reach using classical methods. The presented methodology is accompanied by a practical example to determine the performance of various Machine Learning models for different training sizes.

\end{abstract}


\section{Introduction}

Today's reliability standards and customers' expectations set tough
targets for the quality of electronic devices and systems. Among other
reliability threats, transient faults, such as Single-Event Upsets in
sequential/state logic and Single-Event Transients in combinatorial
logic, are known to contribute significantly to the overall failure
rate of the system. Therefore, estimating the Soft-Error Rate of modern 
complex circuits is a challenging and important task. 

Circuits' susceptibility to transient faults/single events is caused
by faults occurring in the circuit's cells and their subsequent
propagation in the system, possibly causing observable effects
(failures) at the system level. The impact of Single-Event Upsets and
Single-Event Transients in individual state and combinatorial cells
has been extensively 
studied~\cite{baumann_radiation-induced_2005, seifert_radiation-induced_2006} 
and for many applications, identified as the leading contributor to the overall failure rate exhibited by the circuit.

\subsection{Objective of Our Methodology}

De-Rating or Vulnerability Factors are a major tool used in today's
Functional Safety analysis. Since it is difficult and computationally
intensive to get accurate per-instance Functional De-Rating
data for the full list of circuit instances by using classical methods,
we propose an approach using Machine Learning algorithms to assist this procedure. Previous works have shown that the masking effects and thus the vulnerability factors can be related to certain characteristics of the circuit, such as circuit structure and signal probability~\cite{wali_low-cost_2017, ruano_methodology_2009, samudrala_selective_2004}. Thus, we assume that machine learning models are able to learn and predict the Functional De-Rating by using such characteristics.
Therefore, an analysis flow is presented which uses Machine Learning models to predict the Functional De-Rating factors of individual flip-flops. A set of features is described to characterise each flip-flop instance in the circuit individually. The flip-flop features are used to train the Machine Learning model in a supervised learning approach. The trained model is able to predict the remaining Functional De-Rating values for the flip-flop instances not used for training. The proposed methodology is validated in a practical example and compared against a full flat statistical fault injection campaign.

\subsection{Organisation of the Paper}

The rest of this paper is organised as follows: Section~\ref{sec:background} summarises the definition of Single-Event Effects and the different de-rating mechanism. Further, regression in context of supervised Machine learning is explained. The proposed methodology and the used feature set is described in section~\ref{sec:methodology}. In section~\ref{sec:results} the proposed method is validate on a practical example by using different Machine Learning models which are compared to each other. Section~\ref{sec:conclusion} summarises this paper and gives concluding remarks as well as prospects for future work.

\section{Background}
\label{sec:background}

\subsection{De-Rating Mechanism}

Erroneous data in one of the memory or logic points of
a circuit can be produced by the propagation of a Single-Event
Transient (SET) or Single-Event Upset (SEU). SETs are the result
of the collection of charge deposited by ionising particles on
combinatorial logic cells. SEUs are the change of the logic state of a
discrete sequential element, such as a latch, a flip flop or a memory
cell. In the data path between flip-flops, four de-rating
mechanisms~\cite{nguyen_systematic_2003, alexandrescu_towards_2012} significantly reduce the impact of SETs and SEUs on the effective error rate.
\begin{LaTeXdescription}
\item [Electrical De-Rating (EDR):] The transient is filtered due to pulse
  narrowing and or an increase of the rise and fall time during its
  propagation. By the time it reaches the end of the path, either it
  has been completely filtered or the voltage transition is below the
  switching threshold.
\item [Temporal De-Rating (TDR):] The erroneous state reaches the
  input of a flip-flop but outside the latching window, thus it is not sampled.
\item [Logical De-Rating (LDR):] The erroneous state is prevented from
  propagating due to the state on another controlling input of a gate
  such as a zero value on an \verb+AND2+ gate.
\item [Functional De-Rating (FDR):] The erroneous state is considered
  at an applicative level. This means even when an SEU/SET does
  propagate (e.g. is not logically or temporally masked), the impact
  at the function of the circuit can vary, and in many cases is
  benign. Thus, considering the faults at an applicative level, the
  de-rating depends on the criteria defining the acceptable behaviour
  of the circuit during the execution of an application and the fault
  classifications (correctable, uncorrectable, not detected by the
  hardware but detected by the software, if a retry is possible, if
  there is a time limit to receive the correct result, etc.)
\end{LaTeXdescription}
These de-rating mechanisms are used to evaluate the probability of the propagation of a fault and are usually estimated by using probabilistic algorithms and simulation based approaches. Thereby, especially the simulation based approaches are very computationally intensive.

\subsection{Supervised Regression with Machine Learning}

Machine Learning (ML) is the concept of a machine learning from examples and making predictions based on its experience, without being explicitly programmed~\cite{alpaydin_introduction_2014}. ML algorithms are usually build upon a mathematical model which uses sample data (also called training data) in order to make predictions or decisions. The machine learning process usually consists of two phases, namely the training or learning phase and the prediction phase. The learning phase can be further grouped in \begin{enumerate*}
    \item supervised learning and
    \item unsupervised learning
\end{enumerate*}. 

Supervised learning algorithms try to model the relationship and dependency between the input features and the target output in such a way that the output values for new data points can be predicted based on the learned relationships. The main tasks of supervised learning models are classification and regression. While classification algorithms are used when the outputs are restricted to a limited set of values, regression algorithms are used when the outputs may have any numerical value within a range.

In contrast to supervised learning, the unsupervised learning models try to find structures in the data set without external labelling or classification. The two main tasks in this type of machine learning methods are clustering and dimensionality reduction. Clustering algorithms are organizing the given data into groups by similarity and dimensionality reduction is compressing the data by reducing redundancy, while maintaining the overall structure.

\bigskip

The objective of this work is to predict continuous Functional De-Rating factors for individual flip-flops with the help of Machine Learning models. Thus, the proposed methodology is based on supervised regression and is presented in the following section.

\section{Methodology}
\label{sec:methodology}

This section presents the proposed methodology to estimate Functional De-Rating factors per flip-flop instance by using Machine Learning regression models. Therefore, the implemented approach is described in detail, including the feature set to characterise each flip-flop instance and the evaluation metrics used to measure the performance of the models.

\subsection{Functional De-Rating Estimation Flow}

\begin{figure}[htbp]
    \centering
    \includegraphics[
        width=1\linewidth
    ]{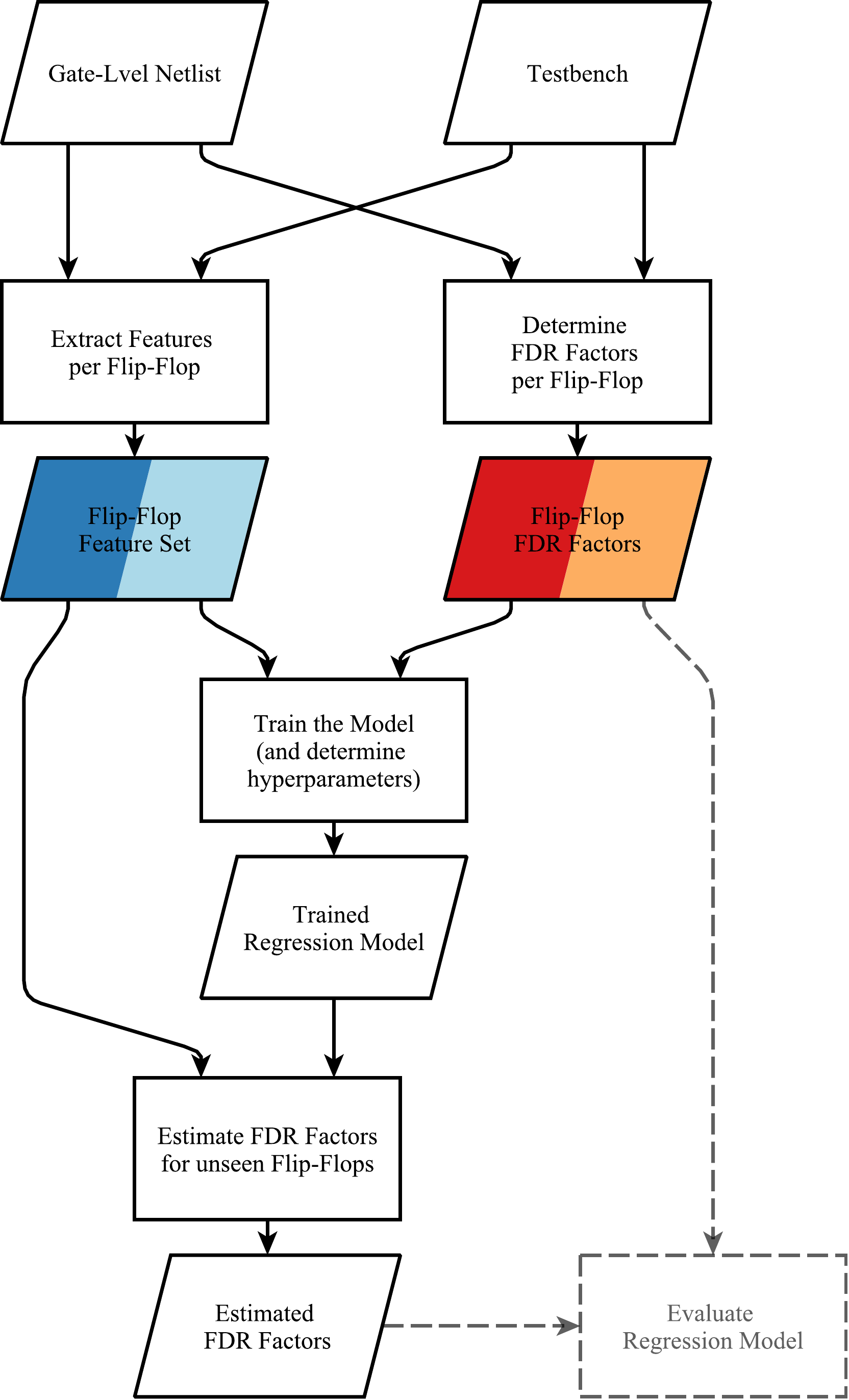}
    \caption{Procedure to Estimate and Evaluate the Functional De-Rating}
    \label{fig:ML_flow}
\end{figure}

The implemented procedure to estimate the Functional De-Rating factors is shown in Fig.~\ref{fig:ML_flow}. It is based on the gate-level netlist of the circuit and a corresponding testbench, which are used to extract the features for each flip-flop in the circuit (the set of flip-flop features is described in section~\ref{sec:ff_features}). Further, they are used to determine the FDR factors for one part of the circuit by using statistical fault injection. The determined FDR factors per flip-flop and the associated flip-flop features form the training data set, used to train the ML model. The size of the training data set is defined by the training size and thus, also defines the number of fault injections to perform.

All ML models have internal model parameters, which are determined during the training process by the ML algorithm. Additionally, most of the ML models also have hyperparameters, which, in contrast to the internal parameters, are manually set before the training process and are not derived by the training algorithm itself. Therefore, several instances of the model
need to be trained and evaluated for different hyperparameters (the used evaluation metrics are described in section~\ref{sec:model_eval_metrics}). A common method to determine the best hyperparameters is to first evaluate the model with randomly selected values for these parameters in a given distribution (random search). Afterwards a more detailed grid search is performed within the region of the values obtained by the random search~\cite{bergstra_random_2012}. The in this way obtained trained model can be used to estimate the FDR values of the remaining flip-flops. 

In this paper we further intend to validate and measure the performance of the proposed approach against a full statistical fault injection campaign (see section~\ref{sec:results}). In order to ensure that the model is not only trained for one particular training and test data set we use the cross-validation technique. Thereby, the model is trained and evaluated against multiple train and test splits of the data. Several subsets, or cross validation folds, of the data set are created and each of the folds is used to train and evaluate a separate model. Thus, we are obtaining a more stable measure of how the model is likely to perform on average, instead of relying only on one single training and test data set~\cite{kohavi_study_1995}. In section~\ref{sec:results} our methodology is evaluated on a practical example and therefore, a ten fold stratified cross validation is used.

\subsection{Flip-Flop Feature Set}
\label{sec:ff_features}

The proposed feature set to characterise each flip-flop instance, combines static elements, such as cell properties, circuit structure and synthesis attributes, as well as dynamic elements, such as signal activity. In order to extract the features describing the circuit structure the the gate-level netlist was converted into a graph representation. Thus, graph algorithms, such as Dijkstra's algorithm to find the shortest path, could be used to extract the features. For each flip-flop $FF_\text{i}$ the following structural features have been extracted.
\begin{LaTeXdescription}
    \item[Flip-Flop Fan-In] This parameter describes how many flip-flops are directly connected (only through combinatorial logic) to the input of the target flip-flop $FF_i$.
    \item[Flip-Flop Fan-Out] This parameter describes to how many flip-flops the target flip-flop's $FF_i$ output is directly connected (only through combinatorial logic).
    \item[Total Flip-Flops from $\bm{FF_i}$] This parameter refers to the number of flip-flops which are influencing the input of the target flip-flop $FF_i$. It represents the total number of flip-flops which are connected to the target flip-flop within the full circuit.
    \item[Total Flip-Flops to $\bm{FF_i}$] This parameter refers to the number of flip-flops which are influenced by the target flip-flop's $FF_i$ output. It represents the total number of flip-flops which are connected to the output of the target flip-flop within the full circuit.
    \item[Connections from Primary Input] This parameter describes how many primary inputs are connected to the target flip-flop's $FF_i$ input.
    \item[Connections to Primary Output] This parameter describes to how many primary outputs the target flip-flop $FF_i$ is connected.
    \item[Proximity from Primary Input] This parameter refers to the proximity of the primary input to the target flip-flop $FF_i$. It represents the number of stages from the connected primary inputs to the target flip-flop. Thereby, the maximum, average and minimum number of stages are considered.
    \item[Proximity to Primary Output] This parameter refers to the proximity of the target flip-flop's $FF_i$ output to the primary output. It represents the number of stages from the target flip-flop to the primary outputs. Thereby, the maximum, average and minimum number of stages are considered.
    \item[Part of Bus] This parameter describes if the target flip-flop $FF_i$ is part of a bus or not.
    \item[Bus Position] If the target flip-flop $FF_i$ is part of a bus, then this parameter represents the position within the bus. It is set to $-1$ if the flip-flop is not part of a bus.
    \item[Bus Length] If the target flip-flop $FF_i$ is part of a bus, this parameter represents the length of the bus. It is set to $0$ if the flip-flop is not part of a bus.
    \item[Connections to constant drivers] This parameter refers to the number of connected constant drivers to the target flip-flop $FF_i$. It represents how many constant drivers are connected to the flip-flop's input within the circuit.
    \item[Has Feedback Loop] This parameter refers to the situation in which the output signal of the target flip-flop $FF_i$ is passed to its input, directly or through several flip-flop stages.
    \item[Depth of Feedback Loop] If the target flip-flop $FF_i$ has a feedback loop, directly or through several flip-flop stages, this parameter describes the minimum number of stages. It is set to $-1$ if there is no feedback loop.
\end{LaTeXdescription}

Further features were extracted which are related to the synthesis of the circuit and were obtained by using Synopsys Design Compiler. The following  synthesis related features have been extracted for each flip-flop~$FF_i$.
\begin{LaTeXdescription}
    \item[Flip-Flop Drive Strength] This parameter describes the drive strength of the target flip-flop $FF_i$ selected by the synthesis tool.
    \item[Combinatorial Fan-In] This parameter describes the number of combinatorial elements connected to the target flip-flop's $FF_i$ input up to the previous flip-flop stage.
    \item[Combinatorial Fan-Out] This parameter describes the number of combinatorial elements which are driven by the target flip-flop's $FF_i$ output up to the next flip-flop stage.
    \item[Combinatorial Path Depth] This parameter describes the depth of the combinatorial stage at the target flip-flop's $FF_i$ output.
\end{LaTeXdescription}

To consider the workload of the circuit, features are required which describe the dynamic behaviour of the flip-flops. Therefore, the signal activity for each flip-flop is extracted. These are obtained by simulating the gate-level netlist with the corresponding testbench and tracing the signal changes at the output of the flip-flops. For each flip-flop $FF_i$ the following dynamic features have been extracted.
\begin{LaTeXdescription}    
    \item[@0] This parameter refers to the time the output of the target flip-flop $FF_i$ is at logical \verb+0+. It represents the time ratio the flip-flop's output has been at \verb+0+ in relation to the total testbench run time.
    \item[@1] This parameter refers to the time the output of the target flip-flop $FF_i$ is at logical \verb+1+. It represents the time ratio the flip-flop's output has been at \verb+1+ in relation to the total testbench run time.
    \item[State Changes] This parameter refers to the number of changes the target flip-flop's $FF_i$ output has performed. It represents the number of changes from \verb+0+ to \verb+1+ and vice versa.
\end{LaTeXdescription}

\subsection{Regression Model Evaluation Metrics}
\label{sec:model_eval_metrics}

The performance of the Machine Learning model is evaluated by using several metrics. In the following description of these metrics, $\hat{y}_i$ is the value of the $i$-th sample predicted by the model and $y_i$ is the corresponding true/expected value.
\begin{LaTeXdescription}
    \item[Mean Absolute Error] The mean absolute error (MAE) describes the
    average absolute difference of the expected values to the predicted values.
    It is calculated over $n_\text{samples}$ by the following equation (values
    closer to zero are better)
    \begin{equation}
        \text{MAE}(y,\hat{y}) = \frac{1}{n_\text{samples}} \sum_{i=1}^{n_\text{samples}} | y_i - \hat{y}_i |
    \end{equation}
    
    \item[Maximum Absolute Error] The maximum absolute error (MAX) describes the
    maximum difference of the expected values to the predicted values. The
    equation
    \begin{equation}
        \text{MAX}(y,\hat{y}) = \max_{i \in [1, n_\text{samples}]} | y_i - \hat{y}_i |
    \end{equation}
    calculates the metrics (values closer to zero are better).
    
    \item[Root Mean Squared Error]
    The root-mean-square error (RMSE) describes the square root of the quadratic
    error of the expected values. In comparison to the mean absolute error the
    root-mean-square error gives a higher weight to larger errors.
    It is calculated over $n_\text{samples}$ by the following equation (values closer to zero are better)
    \begin{equation}
        \text{RMSE}(y,\hat{y}) = \sqrt{ \frac{1}{n_\text{samples}} \sum_{i=1}^{n_\text{samples}} (y_i - \hat{y}_i)^2 }
    \end{equation}
    
    \item[Explained Variance] 
    The Explained Variance (EV) measures the proportion to which a
    model accounts for the variation (dispersion) of a given data set.
    If $\text{Var}(X)$ is the variance, the square of the standard deviation,
    of a random variable~$X$ then the explained variance is calculated as
    follows
    \begin{equation}
        \text{EV}(y,\hat{y}) = 1 - \frac{\text{Var}(y-\hat{y})}{\text{Var}(y)}
    \end{equation}
    The best possible values is $1$ and lower values are worse.
    
    \item[Coefficient of Determination] The coefficient of determination ($R^2$) provides a measure of how well future samples are likely to be predicted by the model. If $\bar{y}$ is the mean of the expected values, the coefficient of determination can be calculated by
    \begin{equation}
        R^2(y,\hat{y}) = 1 - \frac{%
                \sum_{i=1}^{n_\text{samples}} (y_i - \hat{y}_i)^2%
            }{%
                \sum_{i=1}^{n_\text{samples}} (y_i - \bar{y})^2%
            }
    \end{equation}
    and the best possible value is $1$ (lower values are worse).
\end{LaTeXdescription}

\section{Estimating Functional De-Rating Factors by Using Machine Learning}
\label{sec:results}

In this section the presented methodology is evaluated on a practical example. 
Therefore the Ethernet 10GE~MAC Core from OpenCores is
used. This circuit implements the Media Access Control (MAC)
functions for 10\,Gbps operation as defined in the IEEE~802.3ae standard. The 10GE~MAC core has a 10\,Gbps interface
(XGMII TX/RX) to connect it to different types of Ethernet PHYs and
one packet interface to transmit and receive packets to/from the user
logic~\cite{andre_tanguay_10ge_2013}. The circuit consists of control
logic, state machines, FIFOs and memory interfaces. It is implemented
at the Register-Transfer Level (RTL) and is publicly available on
OpenCores.

The corresponding testbench writes several packets to the 10GE~MAC
transmit packet interface. As packet frames become available in the
transmit FIFO, the MAC calculates a CRC and sends them out to the
XGMII transmitter. The XGMII~TX interface is looped-back to the
XGMII~RX interface in the testbench. The frames are thus processed by
the MAC receive engine and stored in the receive FIFO. Eventually, the
testbench reads frames from the packet receive interface and prints
out the results~\cite{andre_tanguay_10ge_2013}. During the simulation
all sent and received packages to and from the core are monitored and 
recorded. This record is used as the golden reference for the fault 
injection campaign.

By synthesising the design using the NanGate FreePDK45 Open Cell Library~\cite{stine_freepdk_2007}, the gate-level netlist was obtained and 1054\,flip-flops have been identified. First, a full flat statistical fault injection campaign was performed to get the Functional De-Rating factors for each flip-flop. Further, the respective features for each flip-flop have been extracted. These values are used to train and evaluate different regression models as described in the previous section.

\subsection{Flat Statistical Fault Injection Campaign}
\label{sec:fi_campaign}

In order to provide an objective measure of the sensitivity of each flip-flop, a flat statistical fault injection campaign was performed on the gate-level netlist. The fault injection mechanism is implemented by inverting the value stored in a flip-flop using a simulator function. The faults are injected at different times during the active phase of the simulation, when packets are sent and received through the user packet interface.

For each of the 1054\,flip-flops 170 fault injection simulations were performed. The simulation run was considered as a functional failure when the final received packages contained payload corruption or the circuit stopped sending or receiving data. Eventually, the Functional De-Rating factor was calculated by the number of simulation runs with a functional failure divided by the number of total simulation runs.

\subsection{FDR Estimation by Using Different Regression Models}

The results of the full statistical fault injection campaign and the extracted flip-flop features are forming the training and test data set, which are used to train and evaluate different Machine Learning models. All models are implemented using Python's scikit-learn Machine Learning framework~\cite{pedregosa_scikit-learn_2011}, with a cross validation fold of 10 and a training size of 50\,\%. Further, the learning curve was determined, which describes the performance of the model for different training sizes.

\subsubsection{Linear Least Squares Regressor}

The Linear Least Squares algorithm fits a linear model which expects the target value to be a linear combination of the input variables. Thereby, the algorithm aims to minimise the residual sum of squares between the observed responses in the training dataset $y$, and the responses predicted by the linear approximation $\hat{y}$.

\begin{figure}[htbp]
    \centering
    \subfloat[%
        Estimation of the example test data fold (training size = 50\%)%
        \label{fig:LinearRegression_example}]{%
        \includegraphics[trim=660 430 120 100, clip, scale=\resfigscale]
            {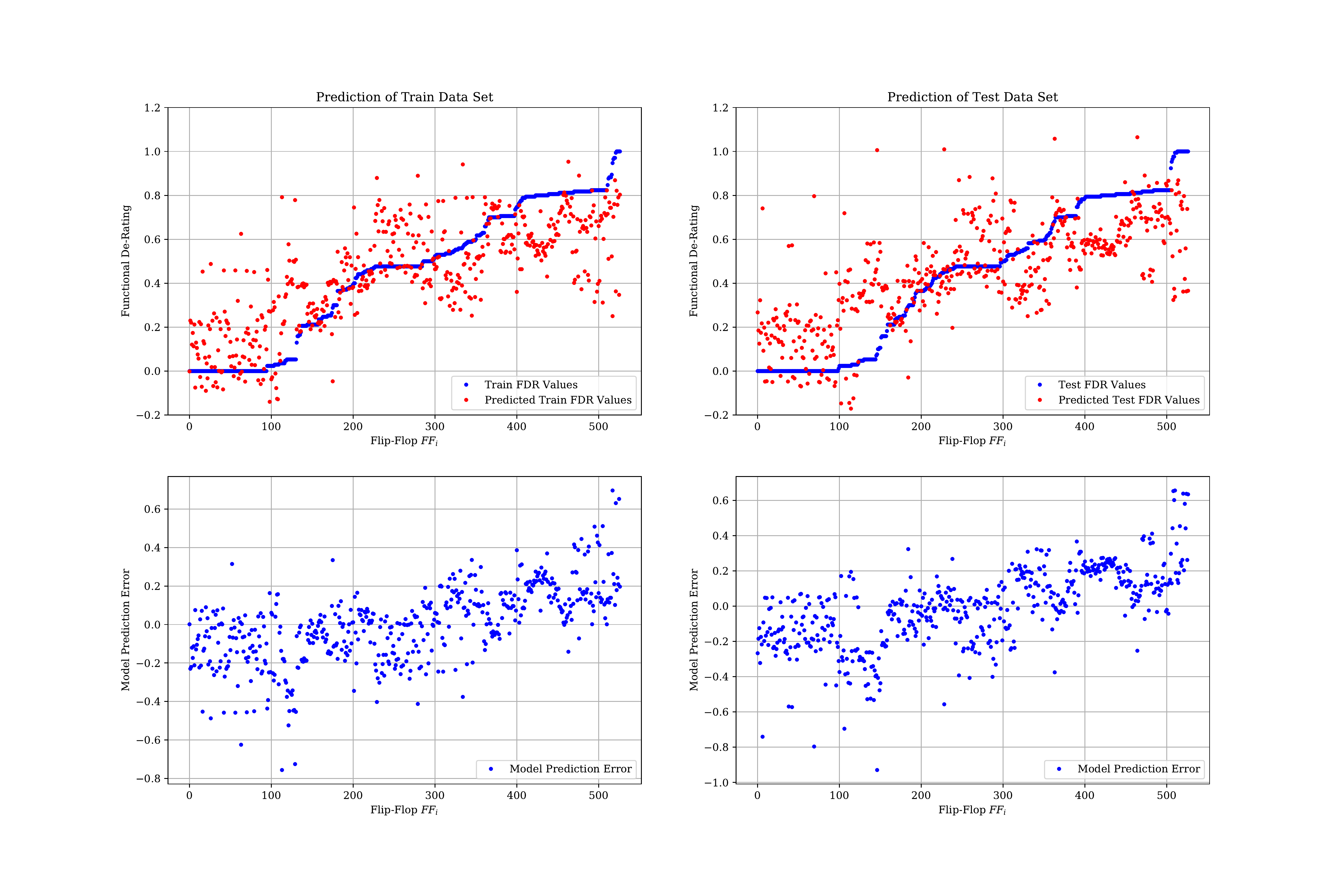}
    }
    
    \subfloat[%
        Learning Curve (cross validation fold = 10)%
        \label{fig:LinearRegression_learn}]{%
        \includegraphics[trim=660 430 120 100, clip, scale=\resfigscale]
            {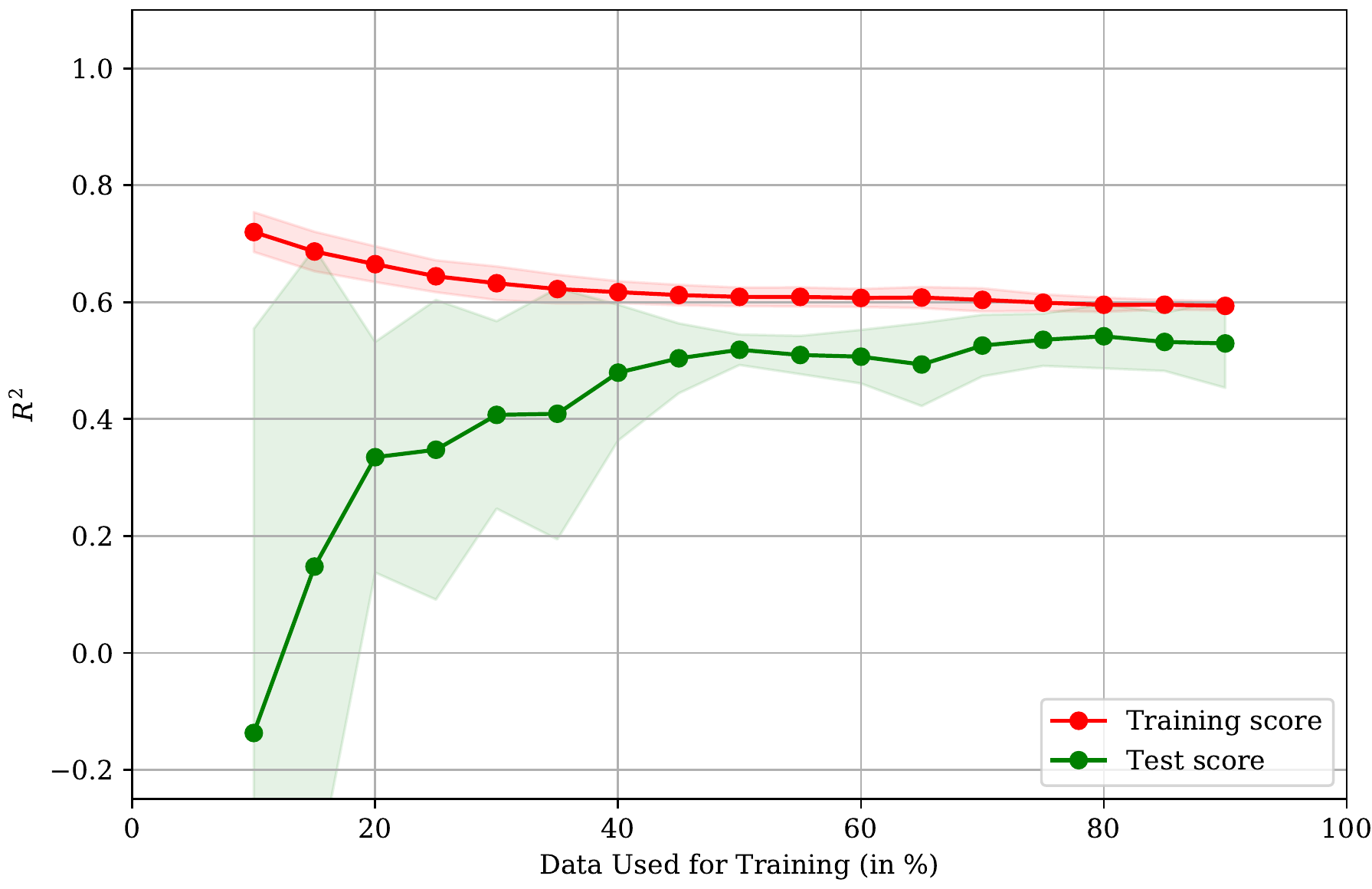}
    }
    \caption{Regression with the Linear Least Squares model}
    \label{fig:LinearRegression}
\end{figure}

The performance of the Linear Least Squares model is given in Table~\ref{tab:estimation_restults}. A regression by using the trained model on an example test data fold is shown in Fig.~\ref{fig:LinearRegression_example} and Fig.~\ref{fig:LinearRegression_learn} shows the learning curve of the model.

\subsubsection{k-Nearest Neighbors Regressor}

The principle behind the $k$-Nearest Neighbor ($k$-NN) regressor is to use feature similarity to predict values of new data points. The new point to predict is assigned a value based on how closely it resembles to the points in the training set. A weighted average of the $k$~nearest neighbors is used to predict the value, where the weight is calculated by the inverse of the distances and the distance itself can be any metric measure, such as the Manhattan or Euclidean distance. Hence, the model defines $k$ and the distance metrics as hyperparameters.

\begin{figure}[htbp]
    \centering
    \subfloat[%
        Estimation of the example test data set (training size = 50\%)%
        \label{fig:knn_example}]{%
        \includegraphics[trim=660 430 120 100, clip, scale=\resfigscale]
            {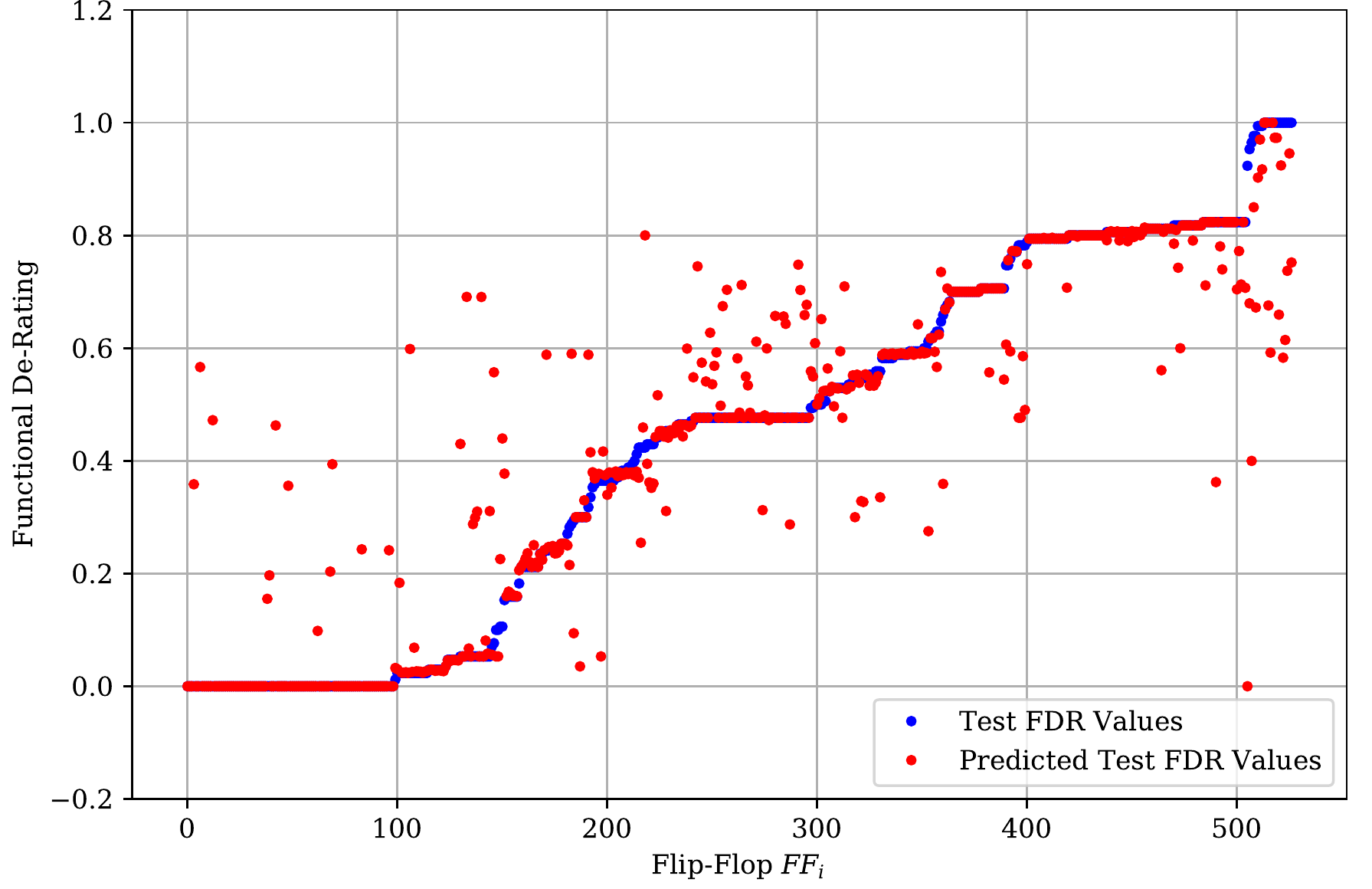}
    }
    
    \subfloat[%
        Learning Curve (cross validation fold = 10)%
        \label{fig:knn_learn}]{%
        \includegraphics[trim=660 430 120 100, clip, scale=\resfigscale]
            {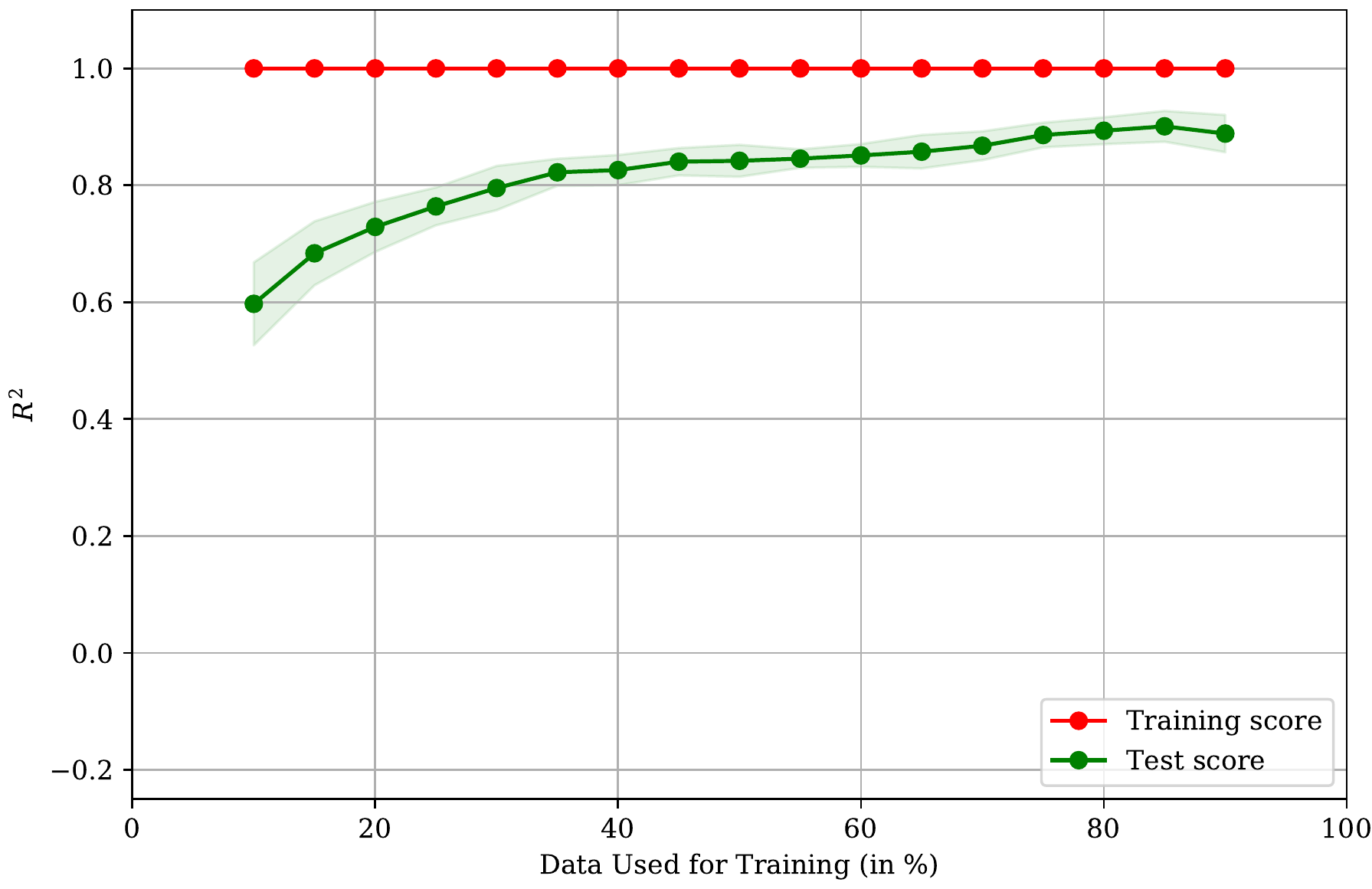}
    }
    \caption{Regression with the k-Nearest Neighbors model}
    \label{fig:knn}
\end{figure}

The best values for the hyperparameters $k$ and the distance metrics, found during the training phase by using random and grid search, are $k=3$ and the Manhattan distance. The resulting performance of the $k$-NN model is listed in Table~\ref{tab:estimation_restults}. A regression with the trained model on the example test data fold is shown in Fig.~\ref{fig:knn_example} and Fig.~\ref{fig:knn_learn} shows the learning curve for the $k$-NN model.

\subsubsection{Support Vector Regression with RBF Kernel}

The goal of the Support Vector Regression (SVR) is to find a function that deviates from the target value by a value not greater than $\varepsilon$ for each training point, and at the same time is as flat as possible. The SVR can be extended to use nonlinear kernel functions, which perform a transformation of the input values and map them to a higher dimensional space. This is useful for regression problems which cannot adequately be described by linear models. In this paper the Radial Basis Function (RBF) was used as kernel function. The model defines several hyperparameters, such as the penalty factor~$C$, the size of the $\varepsilon$-tube, and $\gamma$ to control the RBF kernel function.

\begin{figure}[htbp]
    \centering
    \subfloat[%
        Estimation of the example test data set (traning size = 50\,\%)%
        \label{fig:svr_example}]{%
        \includegraphics[trim=660 430 120 100, clip, scale=\resfigscale]
            {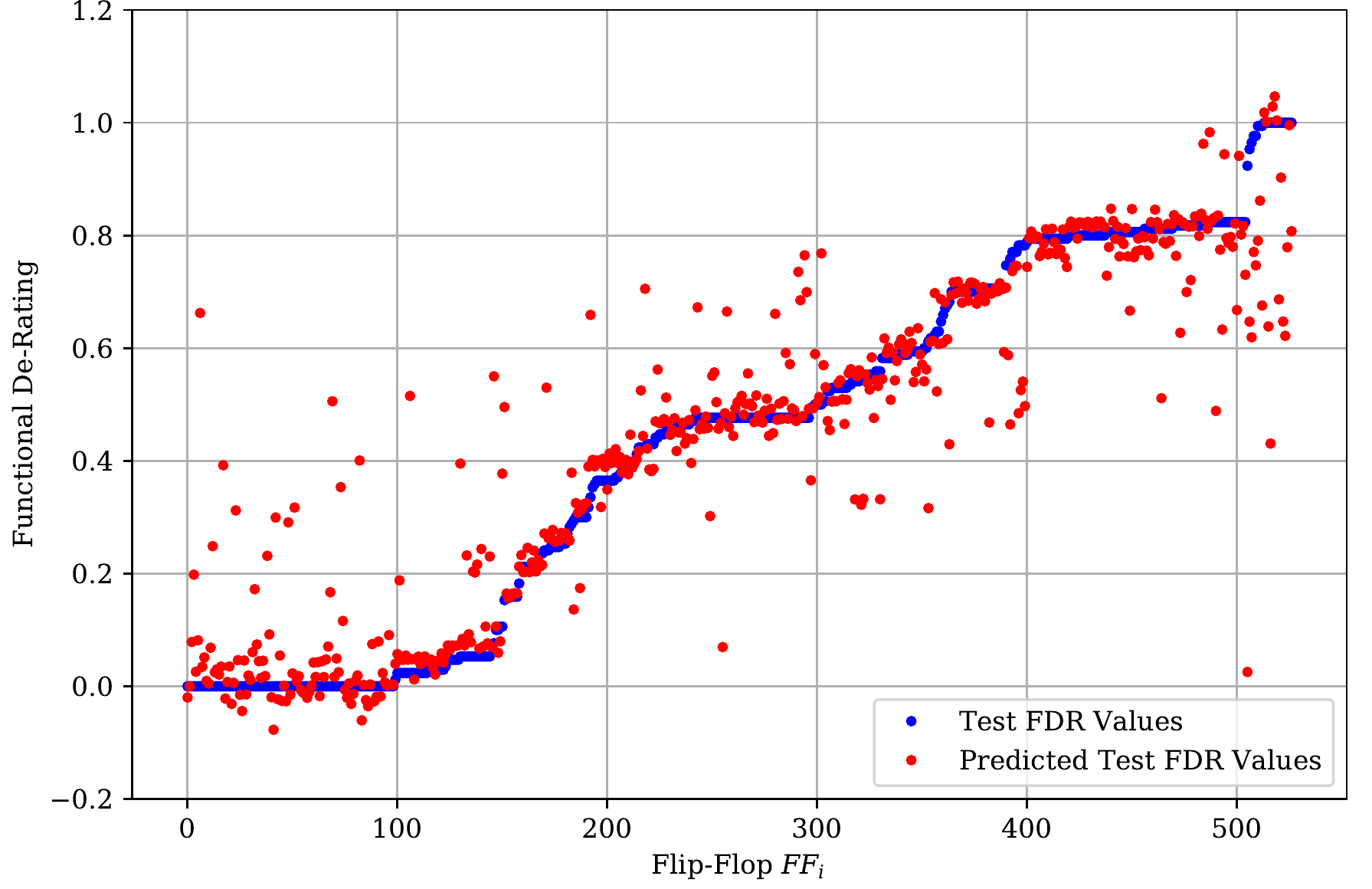}
    }
    
    \subfloat[%
        Learning Curve (cross validation fold = 10)%
        \label{fig:svr_learn}]{%
        \includegraphics[trim=660 430 120 100, clip, scale=\resfigscale]
            {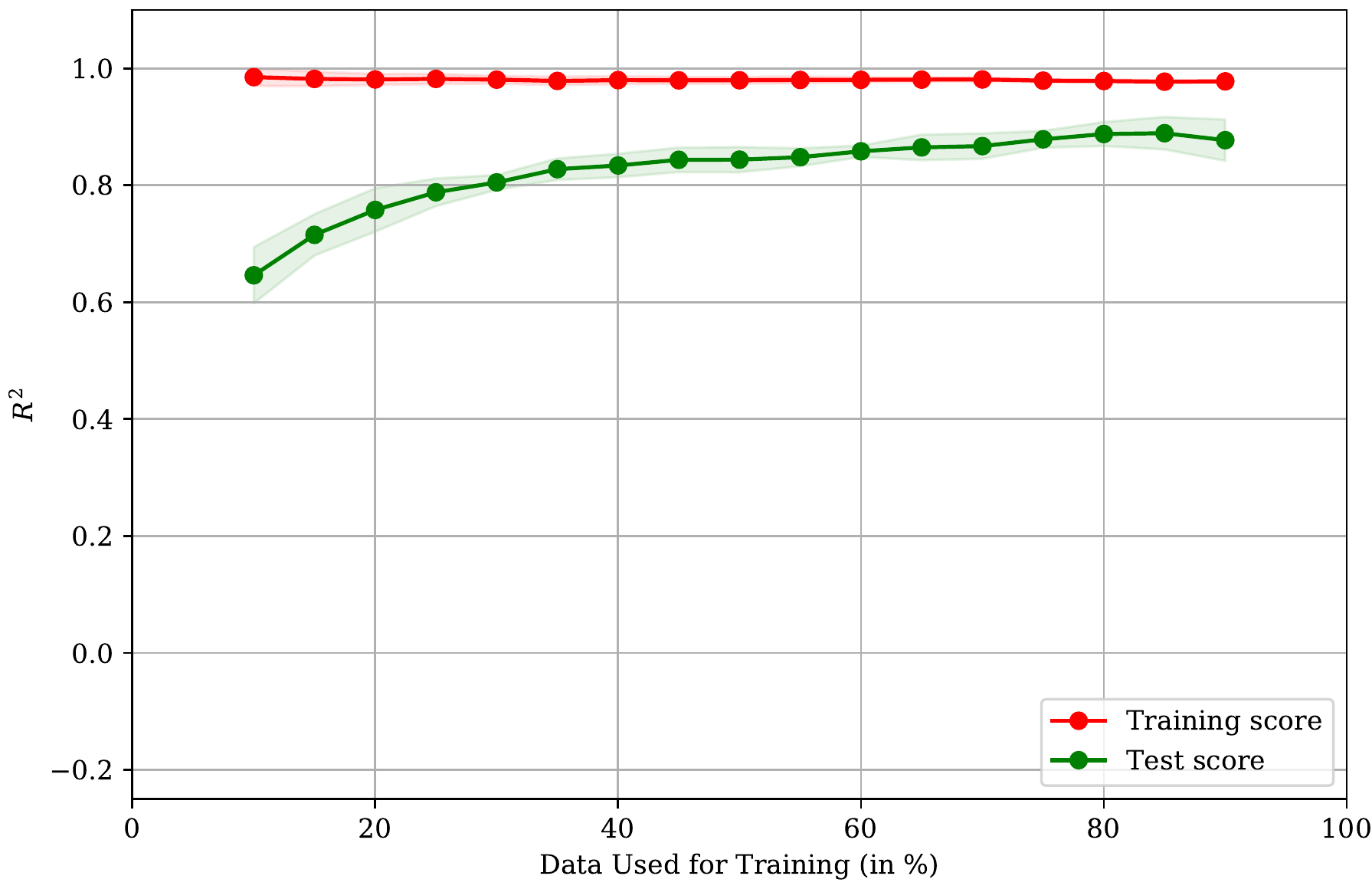}
    }
    \caption{Regression with the Support Vector Regressor with RBF Kernel}
    \label{fig:svr}
\end{figure}

The performance of the Support Vector regression is given in Table~\ref{tab:estimation_restults}. The best hyperparameters of the model found during the training by using random and grid search are $C=3.5$, $\gamma=0.055$ and $\varepsilon=0.0.025$. A regression with the trained model on the example test data fold is shown in Fig.~\ref{fig:svr_example} and the learning curve of the model is shown in Fig.~\ref{fig:svr_learn}.

\subsubsection{Comparison}

By comparing the performance of the different models, shown in Table~\ref{tab:estimation_restults}, it can be seen that the Linear Least Squares model is rated the worst. The other models are performing much better which suggests that the extracted features are not linear dependent to the Functional De-Rating factor and the problem can not be solved with linear models. Further, all models show that the performance does not improve significantly with training sizes higher than 50\,\%. This means, by using the proposed method, the cost for a fault injection campaign can be reduced by half. Further, the learning curves show a more aggressive optimisation can be achieved (a cost reduction up-to $5\times$) in exchange of a slight reduction in accuracy ($< 10\,\%$).

\begin{table}[htbp]
    \centering
    \caption{Performance Results for Different Regression Models \newline
    (Cross Validation = 10, Training Size = 50\,\%)}
    \label{tab:estimation_restults}
    \begin{tabular}{lcccccc}
    \toprule
        \heading{Model} &
            MAE & MAX & RMSE & EV & $R^2$ \\
    \midrule
        Linear Least Sqares &
            0.165 & 0.944 & 0.218 & 0.520 & 0.519 \\
        k-NN &
            0.050 & 0.907 & 0.124 &	0.843 & 0.842 \\
        SVR w/ RBF Kernel &
            0.063 & 0.849 & 0.124 & 0.845 & 0.844 \\
    \bottomrule
    \end{tabular}
\end{table}

\section{Conclusion and Future Work}
\label{sec:conclusion}

In this paper we proposed a new methodology to assist the Functional Failure analysis by using Machine Learning models. The methodology helps to reduce the cost of computing the Functional De-Rating factors of sequential logic of a circuit. Specifically, the methodology allows the computation of factors per individual instances, which is particularly difficult to obtain using state-of-the-art approaches such as clustering, selective fault simulation or fault universe compaction techniques. Therefore, we propose a feature set to describe the individual flip-flops and an estimation flow to train and evaluate the Machine Learning model.

The methodology was evaluated in a practical example. The comparison of the performance of different models has shown that the linear model is not able to fit the problem. Further, the practical example has shown that training sizes of 20\% to 50\% provides appropriate performance, which means that the cost for a classical statistical fault injection campaign could be reduced by 2 up to 5 times.

The focus for future work should lie on evaluating further non-linear models, such as Decision Tree Regressor, Multi-Layer Perception Neural Networks, or using boosting algorithms. Additionally, further features should be considered to improve the overall performance of the models. However, also a dimension reduction should be taken into account in order to avoid the curse of dimensionality and the value of each feature needs to be evaluated separately~\cite{trunk_problem_1979}.

\bibliographystyle{IEEEtran}
\bibliography{IEEEabrv,bib/SELSE2019.bib}

\end{document}